\documentclass[letterpaper, 10 pt, conference]{ieeeconf}
\IEEEoverridecommandlockouts
\usepackage{cite}
\usepackage{amsmath,amssymb,amsfonts}
\usepackage{graphicx}
\usepackage{textcomp}
\usepackage{xcolor}

\usepackage{multirow}
\usepackage{caption}  
\usepackage{bm}
\usepackage{array}        
\usepackage{diagbox}      
\usepackage{xcolor}       
\usepackage{booktabs}     
\usepackage{colortbl}     
\usepackage{algorithm}
\usepackage{algpseudocode}
\usepackage{pifont}
\usepackage{makecell}
\usepackage[colorlinks=true, linkcolor=blue, urlcolor=blue, citecolor=blue]{hyperref}

\def\BibTeX{{\rm B\kern-.05em{\sc i\kern-.025em b}\kern-.08em
    T\kern-.1667em\lower.7ex\hbox{E}\kern-.125emX}}
\begin{document}

\title{DyDexHandover: Human-like Dynamic Dexterous Bimanual Handover using RGB-only Perception
\thanks{All authors are with Xiaomi Robotics. \\Corresponding author: Shuaijun Wang, email:wukongwoong@gmail.com.}
\author{Haoran Zhou, Yangwei You, Shuaijun Wang}
}

\maketitle

\begin{abstract}

Dynamic in-air handover is a fundamental challenge for dual-arm robots, requiring accurate perception, precise coordination, and natural motion. Prior methods often rely on dynamics models, strong priors, or depth sensing, limiting generalization and naturalness. We present DyDexHandover, a novel framework that employs multi-agent reinforcement learning to train an end-to-end RGB-based policy for bimanual object throwing and catching. To achieve more human-like behavior, the throwing policy is guided by a human-policy regularization scheme, encouraging fluid and natural motion, and enhancing the policy’s generalization capability. A dual-arm simulation environment was built in Isaac Sim for experimental evaluation. DyDexHandover achieves nearly 99\% success on training objects and 75\% on unseen objects, while generating human-like throwing and catching behaviors. To our knowledge, it is the first method to realize dual-arm in-air handover using only raw RGB perception. Our project website is available at \href{https://sites.google.com/view/dydexhandover}{https://sites.google.com/view/dydexhandover}
\end{abstract}


\section{Introduction}

\begin{figure}[htbp]
    \centering
    \includegraphics[width=3.0in]{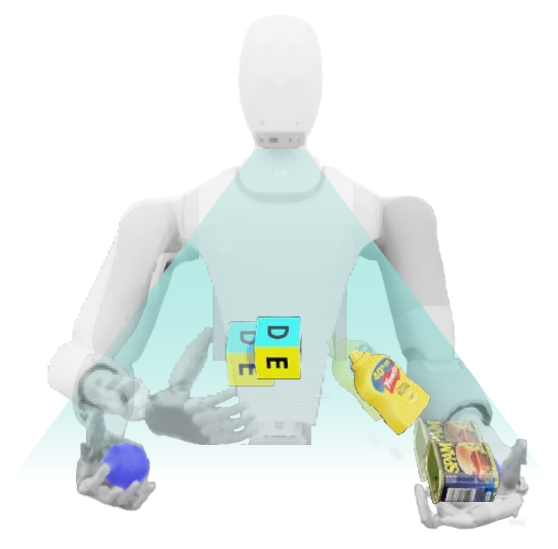}
    \caption{We present \textbf{DyDexHandover}, a framework for dynamic bimanual handover, achieving human-like throwing and catching using RGB-only perception.}
    \label{fig:teaser}
\end{figure}

For humans, the ability to throw and catch with both hands is an essential skill in daily life, significantly enhancing the efficiency of object transfer. Inspired by this, bimanual robotic equipped with dexterous hands can emulate human manipulation behaviors, facilitating the acquisition of dexterous manipulation skills \cite{shao2024bimanual,wang2024learning,zhou2024learning}. As shown in Fig.~\ref{fig:teaser}, once robots possess the capability to perform bimanual throwing and catching, they can not only improve the efficiency of transfer tasks but also extend their operational workspace beyond their physical motion limits \cite{huang2023dynamic,wang2017generic}. More importantly, object transfer via throwing effectively avoids potential collisions that may occur during direct hand-to-hand exchange, providing a contact-free and safer solution for bimanual robotic collaboration.

At present, data-driven methods have been widely applied to manipulation tasks \cite{ wang2023learning,chen2024object,wang2022learning, zhou2025intelligent}. However, most existing works primarily focus on interactions with static objects, such as grasping or transporting, while dynamic tasks like object throwing and catching have received relatively little attention. The main challenges of the throwing and catching task include: (1) the high-dimensional action and observation spaces, which increase the difficulty of learning robust policies; (2) the stringent requirements on response speed and motion precision; and (3) the need for real-time hand–arm and arm–arm coordination. 

Due to the aforementioned limitations, it is difficult to collect effective demonstrations to support training for imitation learning (IL).
Existing RL-based works \cite{kober2012learning, chen2022towards,lan2023dexcatch,zhan2024safe} rely on privileged information or observations containing depth, such as RGBD or point clouds. On the other hand, these works typically employ collaborative manipulators with constrained joints, resulting in motions that are less natural than human throwing and catching strategies.
Inspired by this, we directly employ a pre-trained visual encoder to extract features from RGB images as observation inputs, thereby further enhancing the policy’s generalization ability to unseen objects. Moreover, since the throwing behavior directly affects the predictability of the object’s trajectory and the difficulty of catching it, employing a human-like throwing strategy that propels the object along a higher-curvature trajectory can enhance the overall task robustness, which is of significant importance in throwing and catching tasks. Considering this, we first collect a small number of human throwing demonstrations to train a human policy, which is then used to guide the agent’s throwing behavior during training. By learning a human-like thrower, the policy can achieve improved generalization and robustness.

In this work, we propose DyDexHandover, a novel framework for learning human-like bimanual throwing and catching skill using multi-agent reinforcement learning. 
We constructed the training environment in Isaac Sim. The humanoid robot has a dual-arm configuration, with each arm having 7-DOFs and a 11-DOFs (6 active) dexterous hand at the end. Both hands and arms will need to operate collaboratively at the same time.

In summary, the main contributions of this study include:

1) \textbf{Human-like throwing and catching skill.}
We leverage a pretrained human policy as a regularization term to guide and constrain the thrower’s learning, mitigate the lack of naturalness in pure RL policies, and improve its generalization for unseen objects.

2) \textbf{End-to-end RGB-based policy training.}
We train the policy with MAPPO algorithm, observing via a head-mounted monocular RGB camera and using an attention-based visual encoder for end-to-end training and deployment. The policy achieves 99\% success on 9 training objects and 75\% on 9 unseen objects.

3) \textbf{Dynamic bimanual cooperation on humanoid.}
We enable a humanoid robot to perform human-like, unconstrained, and dynamic dual arm-hand cooperation in simulation.

\section{Related Work}

\subsection{Multi-Agent Reinforcement Learning}
Multi-agent reinforcement learning (MARL) extends reinforcement learning (RL) to multi-agent systems, enhancing inter-agent cooperation, and improving generalization in unstructured environments \cite{huang2023dynamic}. Multi-Agent Proximal Policy Optimization (MAPPO) \cite{yu2022surprising} further extends the stability and efficiency of PPO to multi-agent settings, effectively addressing non-stationarity among agents through a centralized training with decentralized execution (CTDE) architecture.
In the human brain, each arm is primarily controlled by the contralateral motor cortex \cite{gordon2023somato}, while the corpus callosum facilitates interhemispheric communication to achieve highly coordinated bimanual movements \cite{fitzpatrick2001primary}. Inspired by this mechanism, a dual-arm robot can be modeled as a two-agent multi-agent system. Similarly, Guohui Ding et al. proposed a game-theoretic MARL algorithm that updates Q-values based on Nash equilibrium in a bimatrix Q-game, effectively accomplishing dual-arm collaborative manipulation tasks \cite{ding2020distributed}. Likewise, Luyu Liu et al. combined Hindsight Experience Replay (HER) with MADDPG and introduced a “reward cooperation, penalize competition” mechanism during training, which alleviated self-collision issues in bimanual coordination \cite{liu2021collaborative}.

\subsection{Bimanual Dexterous Manipulation}

In recent years, with increasing attention on dexterous manipulation, data-driven methods have become the main approach used by researchers in this field, achieving high success rates in tasks such as pick-place and in-hand manipulation \cite{lin2025sim,qin2023dexpoint,wu2023learning,van2024geometric}. Similar to our approach, Rajeswaran et al. \cite{rajeswaran2017learning} incorporated a small number of human demonstrations into the RL process to guide policy learning, which not only made the robot’s motions more natural but also improved the robustness of the learned policy. However, for object throwing and catching tasks, coordinated dual-arm manipulation is required, which increases the dimensionality of both the action and observation spaces and poses additional challenges for policy learning. Some prior works have addressed dual-hand dexterous manipulation tasks by combining RL and IL \cite{lin2025sim,wang2024dexcap}. Unlike these approaches that rely on RGB-D or point cloud observations, our method uses only RGB images for perception, which simplifies the input representation and reduces computational complexity, enabling efficient training in simulation.

\subsection{Dynamic Throwing and Catching}
Compared to static manipulation tasks such as pick-and-place, dynamic manipulation tasks involve more complex environmental variations, such as throwing an object and catching it \cite{zhang2025catch,zeng2020tossingbot,werner2024dynamic}. These tasks require robots to quickly perceive the state of objects, and adjust their actions in real time. When the robot is required to perform throwing and catching simultaneously, multi-arm coordination, motion synchronization, and real-time adaptation to environmental changes become key challenges\cite{kober2012learning, chen2022towards,lan2023dexcatch,zhan2024safe,kim2025learning}.
For dynamic handover tasks, Yuanpei Chen et al. employed MARL algorithms to achieve human-level performance in bimanual collaboration under varying levels of difficulty \cite{chen2022towards}. To enable policy deployment in the real world, Binghao Huang et al. further integrated a trajectory prediction model to estimate object landing points, successfully training dual-arm robots to perform throwing and catching tasks \cite{huang2023dynamic}. These works demonstrate that MARL can significantly enhance the performance of bimanual cooperative manipulation. However, although these methods perform well during training, they exhibit limited generalization to unseen objects. In our work, we introduce a human policy regularization approach to learn human-like strategies, thereby improving the policy’s generalization capability.

\begin{figure*}[htbp]
    \centering
    \includegraphics[width=6.5in]{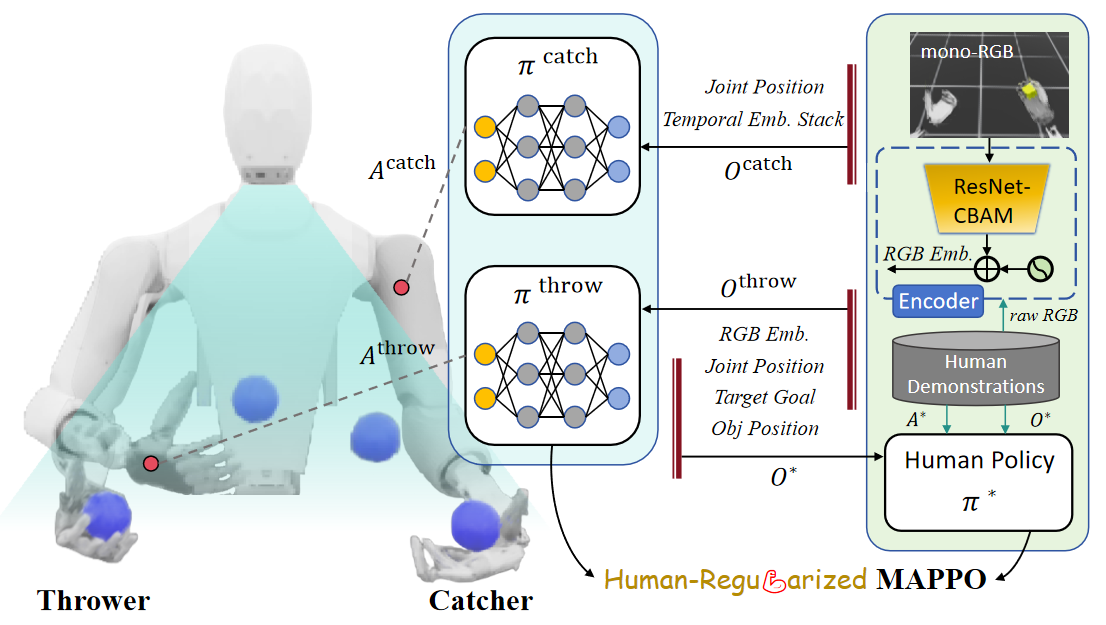}
    \caption{Overview of the proposed framework. We employ the upper body of the M92C humanoid, which features 18-DOFs (7 for the arm and 11 for the hand) on each side. Before RL training, human demonstrations collected via teleoperation are used to pre-train the visual encoder and human policy (right). Subsequently, policies are trained with the MAPPO algorithm, where the human policy is incorporated as a regularization term to guide the training process.}
    \label{fig:framework}
\end{figure*}

\section{Method}
In the throwing and catching task, one hand is responsible for throwing the object while the other hand is responsible for catching it. Both hands share aligned objectives, forming a fully cooperative relationship. Therefore, the humanoid can be regarded as a multi-agent system composed of two agents: the \textbf{thrower} and the \textbf{catcher}. As shown in Fig. \ref{fig:framework}, we train and deploy policies under the CTDE architecture. Each agent observes local information from the environment, executes actions within its own action space, and receives feedback through a shared reward. This modeling approach facilitates the design of strategies that fully exploit dual-arm cooperation, improving the overall efficiency and success rate of the task.

\subsection{Problem Definition}\label{AA}
This work focuses on enabling object throwing and catching tasks on a humanoid robot equipped with dexterous hands through MARL. To achieve this, we formulate the problem as a Decentralized Partially Observable Markov decision process (Dec-POMDP).
And we model the Dec-POMDP for this task, including the design of the action space, state space, and reward function.

\subsubsection{\textbf{Action Space}}
Humans typically perform object throwing and catching by controlling arm velocities in the joint space. To mimic this behavior, we construct the action space of both agents using joint quantities, denoted as:
\begin{equation}
A^i=(\mathbf{\dot{q}}_{d,\text{arm}}^i,\mathbf{q}_{d,\text{hand}}^i),i\in(\text{throw},\text{catch})
\end{equation}
where the 7-DOFs of the arm are controlled via desired joint velocities, denoted as $\mathbf{\dot{q}}_{d,\text{arm}}^i\in\mathbb{R}^7$; while the 6 actuated DOFs of the hand are controlled via desired joint positions, denoted as $\mathbf{q}_{d,\text{hand}}^i\in\mathbb{R}^6$.

\subsubsection{\textbf{State Space}}
In the CTDE framework, the critic receives the global state, while each policy only observes local states. 
In our task, both agents observe the target throwing position 
\( \mathbf{p}_{\text{target}}\in\mathbb{R}^3 \), the joint positions 
\( \mathbf{q}^{\text{throw}},\mathbf{q}^{\text{catch}}\in\mathbb{R}^{13}\) and the visual features \( f\in\mathbb{R}^{8} \) extracted from RGB images $\mathbf{I}$. 
The thrower requires the current visual features, while the catcher requires the past trajectory, define as the temporal features \( \tau_{f} = \{f_{t}, f_{t-1}, \cdots, f_{t-5}\} \), to implicitly capture additional velocity-related features.
Thus, the local state spaces for the two agents are:
\begin{equation}
\mathbf{O}^{\text{throw}} = \{\mathbf{q}^{\text{throw}}, \mathbf{p}_{\text{target}}, f\},
\end{equation}
\begin{equation}
\mathbf{O}^{\text{catch}} = \{\mathbf{q}^{\text{catch}}, \mathbf{p}_{\text{target}}, \tau_{f}\}.
\end{equation}

Since the throwing and catching task involves highly dynamic motions, the joint velocities 
\( \mathbf{\dot{q}}^{\text{throw}},\mathbf{\dot{q}}^{\text{catch}}\in\mathbb{R}^{13} \) play a crucial role in policy learning. 
In addition, the object position \( \mathbf{p}_{\text{obj}}\in\mathbb{R}^3 \) and linear velocity 
\( \mathbf{v}_{\text{obj}}\in\mathbb{R}^3 \) provide complementary information. We can obtain this additional privileged information from the simulation. 
The global state space is defined as:
\begin{equation}
\mathbf{S} = \{\mathbf{O}^{\text{throw}}, \mathbf{O}^{\text{catch}}, \mathbf{\dot{q}}^{\text{throw}},\mathbf{\dot{q}}^{\text{catch}},
\mathbf{p}_{\text{obj}}, \mathbf{v}_{\text{obj}} \}
\end{equation}

\subsubsection{\textbf{Reward Design}}

Given the distances $l_{\text{target}}$ (object to target), $l_{\text{obj}}$ (object to left palm), $l_{\text{hand}}$ (right palm to left palm), a boolean $f_{\text{contact}}$ indicating left-hand contact with the object, the number of robot joints $n$, and the torque of the joint $\tau$. We design five types of reward terms:

\textit{Distance Reward.}  
Encourage the robot to move the object toward the target point:
\begin{equation}
R_{\text{dist}} = \exp(-10 \cdot l_{\text{target}}).
\end{equation}

\textit{Catching Reward.}  
Encourages the left hand to get closer to the object:
\begin{equation}
R_{\text{obj}} = \exp(-10 \cdot l_{\text{obj}})
\end{equation}

\textit{Contact Reward.}  
Encourage the catcher to make contact with the object:
\begin{equation}
R_{\text{contact}} = \mathbb{I}\!\left(f_{\text{contact}}\right),
\end{equation}

\textit{Action Penalty.}  
To encourage smooth and energy-efficient control commands, we apply a penalty based on the instantaneous joint power:
\begin{equation}
P_{\text{action}} = - \sqrt{\sum_{j=1}^{n} (\mathbf{\tau_j} \, \mathbf{\dot{q}}_j)^2},
\end{equation}

\textit{Hand Proximity Penalty.}  
To prevent the policy from exploiting direct hand-to-hand transfers in the overlapping workspace, we penalize configurations where the two hands are too close:
\begin{equation}
P_{\text{hand}} = -\exp(-10 \cdot l_{\text{hand}}).
\end{equation}

\textit{Overall Reward.}  
The final reward is a weighted sum of the above terms:
\begin{equation}
R = 4 R_{\text{dist}} + 0.5 R_{\text{obj}} + R_{\text{contact}} + 0.0001 P_{\text{action}}+P_{\text{hand}}
\end{equation}

\begin{algorithm}
\caption{Human-Regularized MAPPO}
\label{alg:mappo}
\begin{algorithmic}[1]
\State \textbf{Initialize} parameters \(\theta^i\) for policy \(\pi^i\) and \(\sigma^i\) for critic \(V^i, i \in (\textnormal{throw}, \textnormal{catch})\)
\State \textbf{Set} hyperparameters as specified in Table \ref{tab:hyperparameters}, maximum steps $L_{\text{max}}$, sampling steps $T_\text{max}$
\While{iteration$<L_{\text{max}}$}
    \State // 1. Interact in environments to collect data
    \For{$t = 0$ to $T_\text{max}$}
        \State Obtain local observations $o_{t}^i$.
        \State Each agent selects an action $a_t^i\sim\pi(o_t^i)$
        \State Obtain reward $r_t$ and next state $s_{t+1}$ 
    \EndFor
    \State // 2. Calculate necessary information
    \State Read a segment of trajectory $\tau=\{s, a^i, r\}$ 
    \State Predict values $v^i=V^i(s)$
    \State Calculate hybrid advantages $\hat{A}^i=A_{\text{hybrid}}(s, s_{\text{next}}, a^i)$
    \State $D\cup \{s^i, a^i, v^i, \hat{A}^i \}$
    \State // 3. Optimize network parameters $\theta^i$, $\sigma^i$
    \For{epoch = 1 to $K$}
        \State Shuffle the data in the experience buffer $D$
        \For{$j = 0$ to $\frac{T_\text{max}}{B}-1$}
            \State Randomly sample $B$ mini-batches $D_j$
            \State Calculate and maximize $J(\theta^i)$ with $D_j$
            \State Update parameters $\theta^i$ and $\sigma^i$
        \EndFor
    \EndFor
\EndWhile
\end{algorithmic}
\end{algorithm}

\begin{table}[h]
\centering
\caption{The Hyperparameters of Our Training Algorithm}
\label{tab:hyperparameters}
\begin{tabular}{lrlr}
\toprule
Parameter & Value & Parameter & Value \\
\midrule
Discount factor $\gamma$ & 0.995 & PPO clip range $\varepsilon$ & 0.2 \\
GAE parameter $\lambda$ & 0.95 & Learning rate & $1 \times 10^{-4}$ \\
FC layer dimension & 512 & Number of hidden layers & 3 \\
Activation function & elu & Max episode time & 3(s) \\
Training epochs $K$ & 8 & Batchsize $B$ & 4096 \\
Decimation & 2 & Advantage weights $\beta_1,\beta_2$  & $0.01,0.001$ \\
\bottomrule
\end{tabular}
\end{table}

\subsubsection{\textbf{Failure Mechanism}}

To ensure stable training and evaluation, the task episode is terminated once a failure condition is met. Specifically, the following criteria are used:
\begin{itemize}
    \item \textit{Object falling:} The task is defined as failed once the object’s vertical position falls below the predefined threshold $z_{\text{obj}}<0.1$m.
    \item \textit{Deviation from goal:} If the distance between the left palm and the target goal exceeds $0.4\,\text{m}$, the task is regarded as failed.
    \item \textit{Excessive proximity of the hands:} Due to the partially overlapping workspace of the robot’s two hands, we reset the episode whenever the distance $l_{\text{hand}}<0.1$m to prevent direct hand-to-hand object transfer.
    \item \textit{Unexpected contact:} Any unintended arm contact with the environment or self-collisions triggers a reset.
    \item \textit{Object out of view:} If the object is no longer visible in the camera image, the episode is terminated.
\end{itemize}

These conditions jointly define the failure mechanism of the environment, preventing unstable behaviors and ensuring meaningful policy learning.

\subsection{Pre-traing with Human Demonstration}
To collect human demonstrations, we teleoperate the right arm in simulation to throw objects toward the left arm, and recording the states $(A^{\text{throw}},\mathbf{q}^{\text{throw}}, \mathbf{p}_{\text{target}},\mathbf{p}_{\text{obj}},\mathbf{I})$ into a dataset. The collected data is then used to train the human policy and the visual encoder.

\subsubsection{\textbf{Pretrained Vision Encoder}}

We use a pre-trained CBAM-based ResNet \cite{xiao2021trec} as the vision backbone, as shown in Fig. \ref{fig:framework}, where the Convolutional Block Attention Module (CBAM) is incorporated to enhance feature representation of objects in the image, and it is then connected to a two-layer MLP. The visual encoder takes RGB images $\mathbf{I}$ from the head-mounted camera as input and outputs embeddings $f\in\mathbb{R}^8$. These embeddings are then concatenated with processed proprioceptive information, serve as input to the policy.

To improve the encoder’s performance in this task, we trained it on the collected dataset using the Adam optimizer with a learning rate of $1\times10^{-4}$. The encoder maps the output embeddings into a 3-dimensional space via a linear layer to match the supervision labels. To accurately capture object geometry and spatial features, we use the object’s area ratio $\delta$ and centroid $(x_c, y_c)$ as labels, and define the loss as the MSE loss:

\begin{equation}
L = \| f_o - f_i \|_2^2, \quad f_i = (\delta, x_c, y_c)
\end{equation}

These labels are computed from segmentation masks: $\delta$ is the ratio of object pixels to total pixels, and $(x_c, y_c)$ is obtained via the weighted average of mask coordinates.

\subsubsection{\textbf{Pretrained Human Policy}}

The human policy $\pi^*$ takes $o^*=(\mathbf{q}^{\text{throw}}, \mathbf{p}_{\text{target}},\mathbf{p}_{\text{obj}})$ as input, and outputs the action $a^*$. We used a 3-layer MLP, and trained the network on the collected dataset with MSE loss. The policy was trained using the Adam optimizer with a learning rate of 1e-4. The total number of training epochs was 10k.

\subsection{Policy Training}
When training the thrower and catcher using RL, the thrower requires a significant amount of exploration in the early stages of training, during which the catcher may learn many ineffective behaviors, negatively affecting overall training success. Therefore, optimizing the throwing strategy is crucial for improving training efficiency and policy performance. 
To improve convergence and training stability, we implement two components: (1) human policy regularization; (2) hybrid advantage estimation. The training procedure is summarized in Algorithm \ref{alg:mappo}.

\subsubsection{\textbf{Human Policy Regularization}}
To encourage the throwing policy to approximate human behavior, we employ the Kullback–Leibler (KL) divergence to measure the discrepancy between the human policy distribution and the thrower’s policy distribution, and incorporate this discrepancy as a regularization term in the training objective of the thrower. Since we adopt a stochastic policy, each policy outputs a probability distribution modeled as a Gaussian distribution $\pi\sim\mathcal{N}(\mu,\sigma^2)$. The KL divergence between the two distributions can be expressed as:
\begin{equation}
\begin{aligned}
D_{\mathrm{KL}}\left(\pi^{*} \| \pi^{\text{throw}}\right) 
&= \mathbb{E}_{a^{\text{throw}} \in A^{\text{throw}}}\left[\log \frac{\pi^{*}(a^* \mid o^*)}{\pi^{\text{throw}}(a^{\text{throw}} \mid o^{\text{throw}})}\right] \\
&= \log \frac{\sigma^{\text{throw}}}{\sigma_{*}} 
   + \frac{{\sigma^{*}}^{2} + (\mu^{*} - \mu^{\text{throw}})^{2}}{2{\sigma^{\text{throw}}}^{2}} 
   - \frac{1}{2}
\end{aligned}
\end{equation}

After incorporating human policy regularization, the objective function of the throwing policy is given by:
\begin{equation}
J\left(\theta^{\text {throw }}\right)=\left(1-\lambda_{\text {reg }}\right) \cdot J\left(\theta^{\text {throw }}\right)-\lambda_{\text {reg }} \cdot D_{\mathrm{KL}}\left(\pi^{*} \| \pi^{\text {throw }}\right)
\end{equation}
where $\lambda_{\text{reg}}$ denotes the regularization weight, which balances the degree of constraint imposed by the human policy on the throwing policy. A larger $\lambda_{\text{reg}}$ indicates stronger influence from the human policy, leading the thrower’s behavior to more closely align with that of the human.

\subsubsection{\textbf{Hybrid Advantage Estimation}}
In the MAPPO algorithm, the advantage function directly influences the optimization direction of the policy. To accelerate policy learning during the early training stage and to prevent loss of exploration ability caused by local optima in later stages, we propose a hybrid advantage estimation method, defined as:
\begin{equation}
A_{\text{hybrid}}(s_t, s_{t+1}, a_t) = A_{\mathrm{GAE}}(s_t, a_t) + \beta_1 A_{\mathrm{I}}(s_t, s_{t+1}) + \beta_2 A_N
\end{equation}
where $A_{\mathrm{GAE}}(s_t, a_t)$ denotes the generalized advantage function in MAPPO, 
$A_{\mathrm{I}}(s_t, s_{t+1})$ represents the internal advantage function introduced in \cite{lan2023dexcatch}, $A_N \sim \mathcal{N}(0,1)$ denotes the Gaussian noise, $\beta_1$ and $\beta_2$ are the corresponding weighting hyperparameters.

The introduction of internal advantage $A_{\mathrm{I}}(s_t, s_{t+1}) = \min(V(s_{t+1}) - V(s_t), 0)$ enables the advantage function to place greater emphasis on unfavorable situations, thereby reducing the failure probability during the early training stage. 

In MAPPO, the partial observability of agents and the non-stationarity of the environment exacerbate the estimation bias of the advantage, which can easily lead to policy overfitting. By applying noise $A_N$ into the advantage estimation, the exploration capability of the policies can be enhanced \cite{zhang2024vn}.

\section{Experiments}
To validate the effectiveness of the proposed approach, we conduct experiments in simulation related to policy learning and deployment. The evaluation focuses on: (1) the effectiveness and generalization of DyDexHandover; (2) the impact of different components within the framework; and (3) the effect of varying levels of human policy regularization on task performance.

\begin{table}[!htbp]
\centering
\caption{Randomization Parameters.}
\label{table:randomization}
\begin{tabular}{@{\extracolsep\fill}l l c c}%
    \toprule
    \textbf{Group} & \textbf{Parameter} & \textbf{Distribution}  & \textbf{Operation} \\
    \midrule
    \multirow{2}{*}{\textbf{Object}} 
        & Mass     & \( \sim\mu \)(0.3, 0.5)   & Sampling \\
        & Color    & $\sim \mu$(RGB 0–1) & Sampling \\
    \midrule
    \multirow{4}{*}{\textbf{Robot}} 
        & Joint Stiffness    & \( \sim\mu \)(0.75, 1.5)   & Scaling \\
        & Joint Damping    & \( \sim\mu \)(0.75, 1.5)   & Scaling \\
        & Restitution    & \( \sim\mu \)(-0.04, 0.04)   & Additive \\
        & Friction    & \( \sim\mu \)(-0.04, 0.04)   & Additive \\
    \midrule
    \multirow{2}{*}{\textbf{Observation}} 
        & Noise    & \( \sim\mathcal{N} \)(0.0, 0.02)   & Additive \\
        & Bias    & \( \sim\mathcal{N} \)(0.0, 0.001)   & Additive \\
    \midrule
    \multirow{2}{*}{\textbf{Action}} 
        & Noise    & \( \sim\mathcal{N} \)(0.0, 0.002)   & Additive \\
        & Bias    & \( \sim\mathcal{N} \)(0.0, 0.0001)   & Additive \\
    \midrule
    \multirow{1}{*}{\textbf{Environment}} 
        & Background    & $p=0.3,\ \mathcal{N}(0,1)$ & Additive \\
    \bottomrule
\end{tabular}
\end{table}

\begin{figure}[htbp]
    \centering
    \includegraphics[width=3.5in]{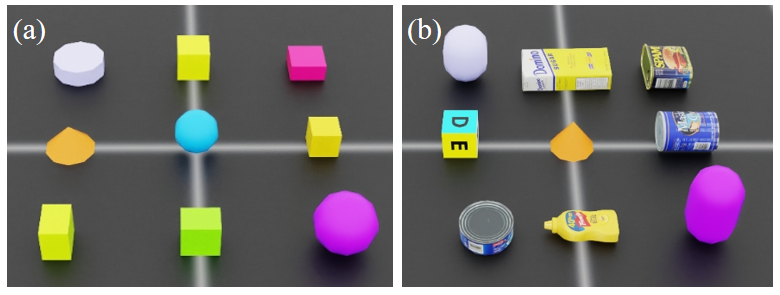}
    \caption{Object sets used in the experiments. (a) Training objects, (b) Unseen objects.}
    \label{fig:objects}
\end{figure}

\begin{figure*}[ht]
    \centering
    \includegraphics[width=7in]{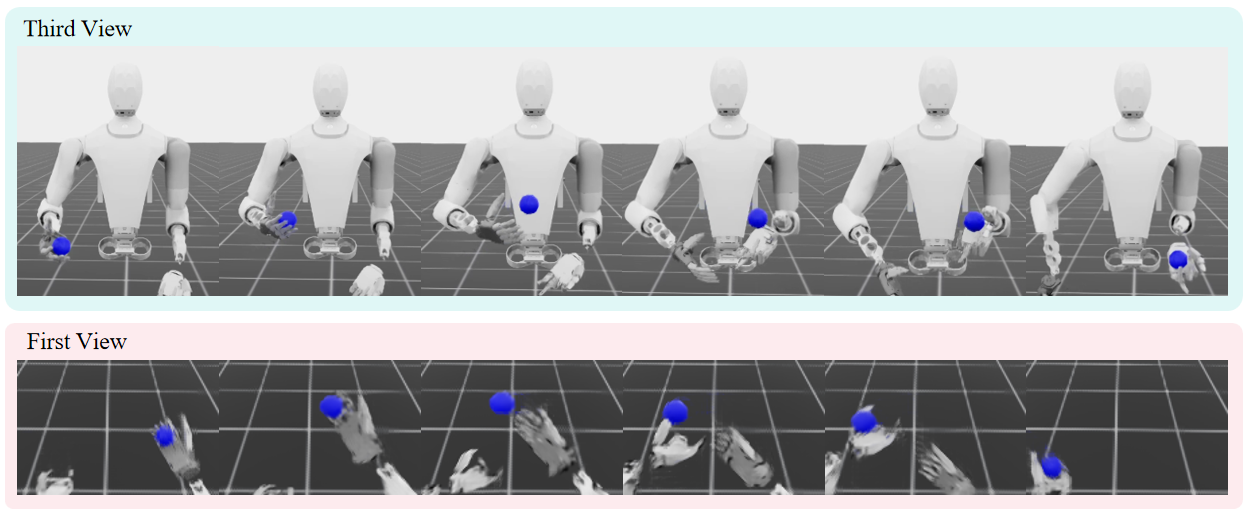}
    \caption{Visualization of the throwing and catching process from two perspectives.}
    \label{fig:process}
\end{figure*}

\begin{table*}[htbp]
\centering
\caption{Evaluation results on predefined metrics.}
\label{tab:evaluation}
\begin{tabular}{lcccccccccc}
    \toprule
    \multicolumn{11}{c}{\textbf{Training objects (9 objects)}} \\
    \midrule
    \textbf{Metric} & Cone & Cube & Big Cube & Cuboid A & Cuboid B & Cuboid C & Sphere & Big Sphere & Cylinder & Avg. \\
    \midrule
    Hit Rate         & 100\% & 100\% & 100\% & 100\% & 100\% & 100\% & 100\% & 100\% & 100\% & \textbf{100\%} \\
    Success Rate     & 100\% & 99\% & 99\% & 99\% & 99\% & 99\% & 100\% & 98\% & 99\% & \textbf{99\%} \\
    \midrule
    \multicolumn{11}{c}{\textbf{Unseen objects (9 objects)}} \\
    \midrule
    \textbf{Metric} & Dex Cube & Chef Can & Sugar Box & Bottle & Fish Can & Meat Can & Long Cone & Capsule & Long Capsule & Avg. \\
    \midrule
    Hit Rate         & 100\% & 93\% & 95\% & 98\% & 83\% & 100\% & 99\% & 98\% & 99\% & \textbf{92\%} \\
    Success Rate     & 97\%  & 69\% & 74\% & 40\% & 73\% & 48\% & 98\% & 92\% & 92\% & \textbf{75\%} \\
    \bottomrule
\end{tabular}
\end{table*}

\begin{figure}[htbp]
    \centering
    \includegraphics[width=3.5in]{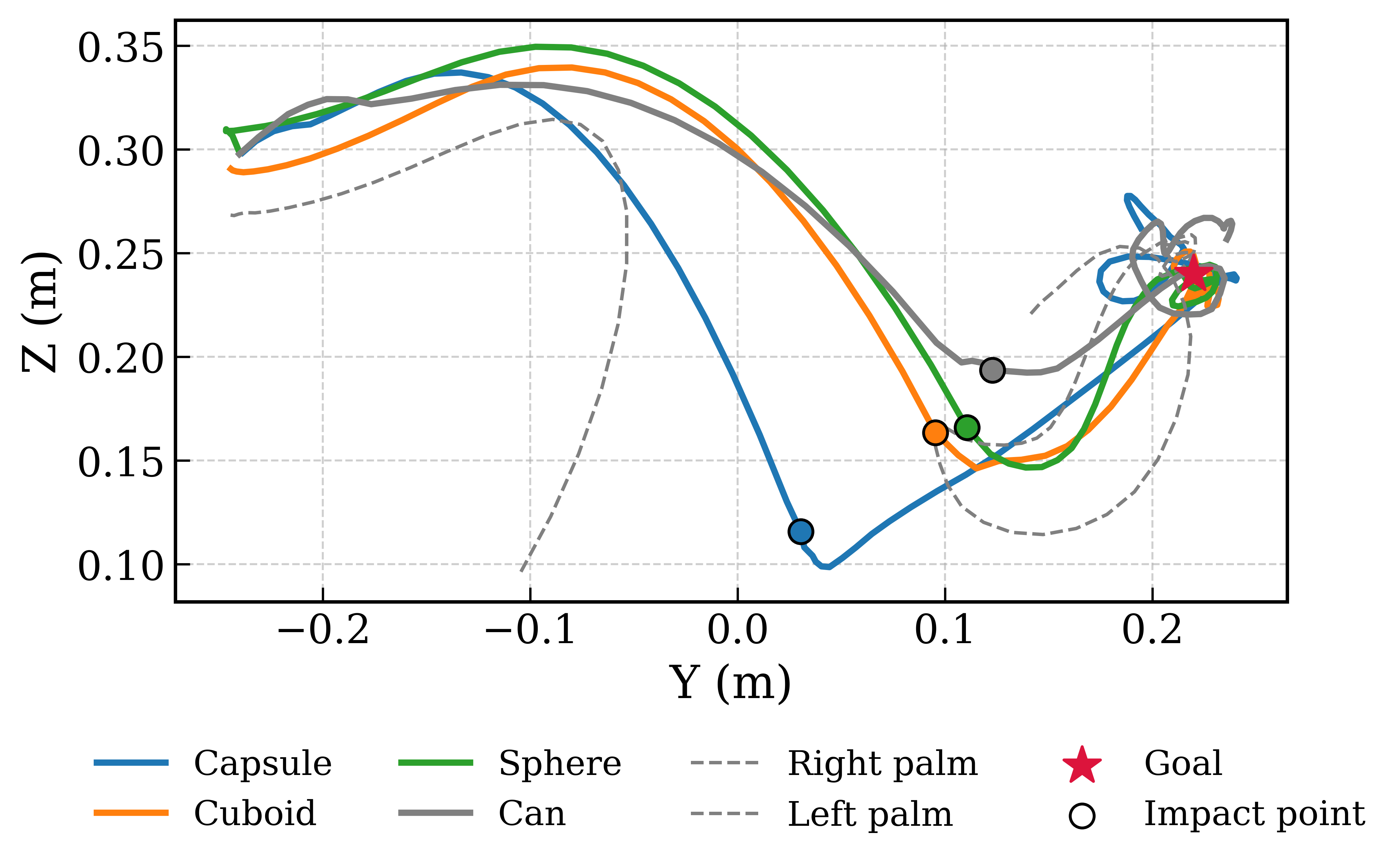}
    \caption{Trajectories of various objects during evaluation.}
    \label{fig:trajectory}
\end{figure}

\subsection{Experimental Details}
In this work, we build a simulated environment for the throwing and catching task based on the NVIDIA Isaac Lab framework \cite{mittal2023orbit}, including the upper body of the M92C-series humanoid robot and the objects used for throwing and catching. A monocular RGB camera is mounted on the robot’s head. During training, each environment randomly samples objects from the training set, as shown in Fig. \ref{fig:objects}. The hardware used in the experiments are as follows: Intel i9-13900K CPU and NVIDIA GeForce RTX 4090 GPU. The physics simulator runs at 120 Hz, with both agents controlled at 60 Hz. To enhance the robustness of the policy, we apply domain randomization during training, as indicated in Table \ref{table:randomization}. 

To evaluate the performance of the policies, we consider several metrics as follow:

1) \textit{Hit Rate.} This metric is defined as the proportion of objects that successfully hit the hand palm of catcher.

2) \textit{Success Rate.} This metric is defined as the proportion of objects that are successfully caught by the catcher and remain in hand until the end of the episode.


\subsection{Evaluation Results}

After training, we evaluate the proposed method on both the object set used during training and an unseen object set. In the test environment, the trained policy is rolled out for 1,000 episodes, with objects randomly sampled from the corresponding set in each episode. The mean values of the evaluation metrics are reported in Table~\ref{tab:evaluation}.
For the training objects, the method achieves an average success rate of 99\%, with 100\% success rate on certain objects. Furthermore, for the unseen objects, our method attains an average success rate of approximately 75\%. These results demonstrate that the proposed method exhibits strong robustness and generalization capability, enabling successful throwing and catching tasks with diverse objects.

Fig. \ref{fig:process} shows the complete throwing and catching process of the humanoid from different viewpoints. It can be observed that the robot performs both throwing and catching in a human-like and natural manner. Fig. \ref{fig:trajectory} provides a more detailed view of the trajectories for different objects, all exhibiting natural parabolic shapes. Notably, when the object hits the catcher, the catcher carries the object along its original trajectory for a short distance to prevent it from bouncing away.

\begin{table*}[!htbp]
\centering
\caption{Different baselines on Bimanual Dexterous Dynamic Handover.}
\label{tab:baseline}
\begin{tabular}{lcccccc}
    \toprule
    \textbf{Algorithm} & \textbf{Obs type} & \textbf{Task type} & \textbf{Method} & \textbf{Train objects} & \textbf{Test objects} & \textbf{Success rate(Train/Test)}   \\
    \midrule
      \text{Bi-DexHands\cite{chen2022towards}}        & State       & Abreast Catch & PPO/MAPPO/HAPPO& 1 & 1 & unknown  \\
      \text{Dynamic Handover\cite{huang2023dynamic}}  & RGB-D        & Underarm Catch & MAPPO& 11 & 14 & 95\%/37\%  \\
      \text{DexCatch\cite{lan2023dexcatch}}           & Point Cloud & Abreast Catch & PPO & 9 & 11 & 72.95\%/77.89\%  \\
      \text{Bimanual Catch\cite{kim2025learning}}     & State       & Bimanual Catch & HAPPO& 15 & 15 & unknown  \\
      \textbf{DyDexHandover (Ours)}                    & RGB         & Abreast Catch & MAPPO& 9 & 9 & 99\%/75\%  \\
    \bottomrule
\end{tabular}
\end{table*}

\begin{figure}[htbp]
    \centering
    \includegraphics[width=3.5in]{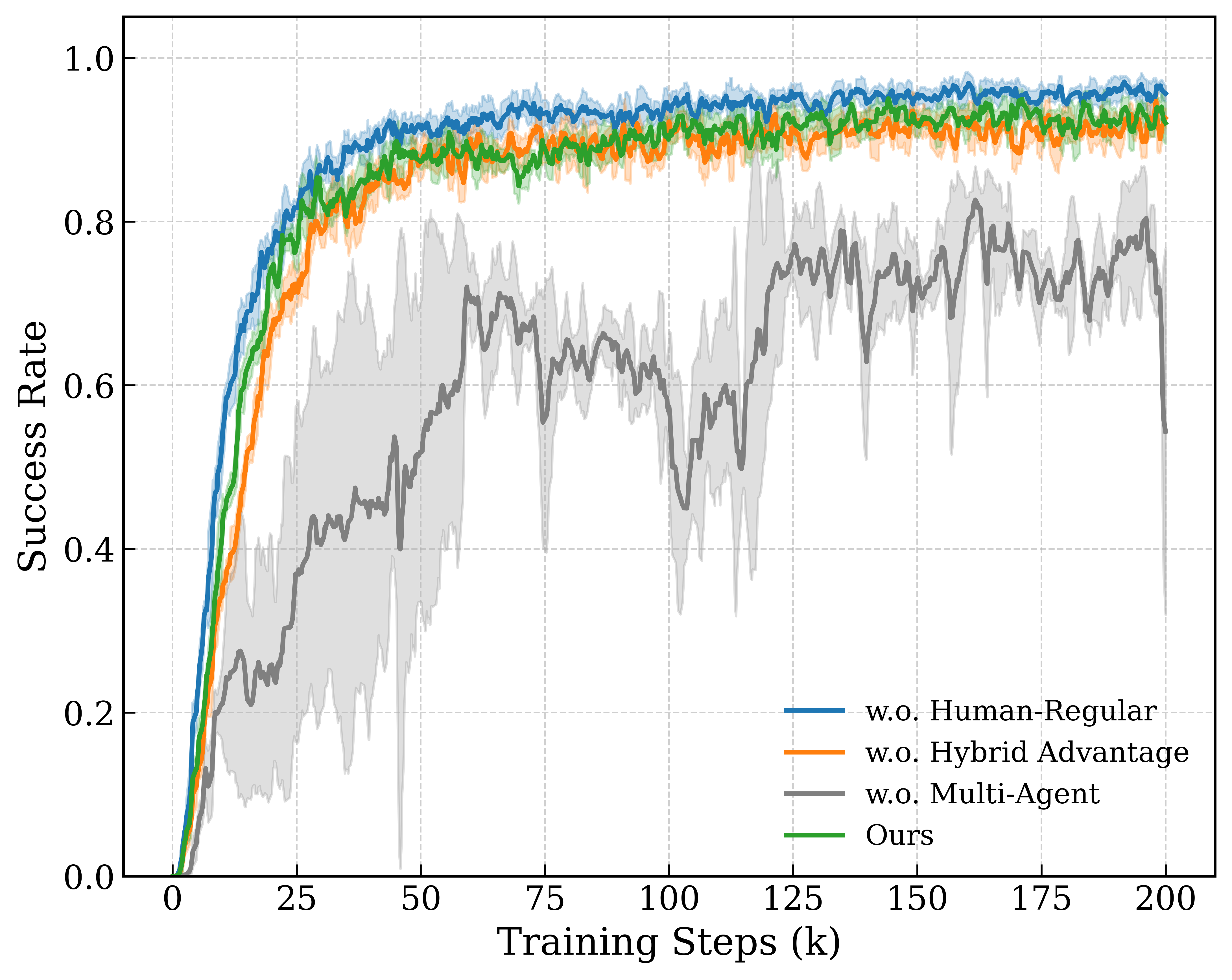}
    \caption{Training curves of average success rate for the ablation study. Each method was trained for 200k steps.}
    \label{fig:ablation}
\end{figure}

\begin{table}[!htbp]
\centering
\caption{Ablation study on predefined metrics.}
\label{tab:ablation}
\begin{tabular}{lcccc}
    \toprule
    & \multicolumn{2}{c}{\textbf{Training objects}} & \multicolumn{2}{c}{\textbf{Unseen objects}} \\
    \cmidrule(lr){2-3} \cmidrule(lr){4-5}
    & Hit rate & Success rate & Hit rate & Success rate \\
    \midrule
      \text{Open-Loop}          & 40\%  & 0\% & X  & X  \\
      \text{w.o. Multi-Agent}   & 89\%  & 79\% & 78\%  & 51\%  \\
      \text{w.o. Human-Reg.}    & 100\% & 97\% & 65\% & 53\%  \\
      \text{w.o. Hybrid Adv.}   & 100\% & 99\% & 86\% & 74\%  \\
      \textbf{Ours}             & 100\% & 99\% & 92\% & 75\%  \\
    \bottomrule
\end{tabular}
\end{table}

\begin{figure}[htbp]
    \centering
    \includegraphics[width=3.5in]{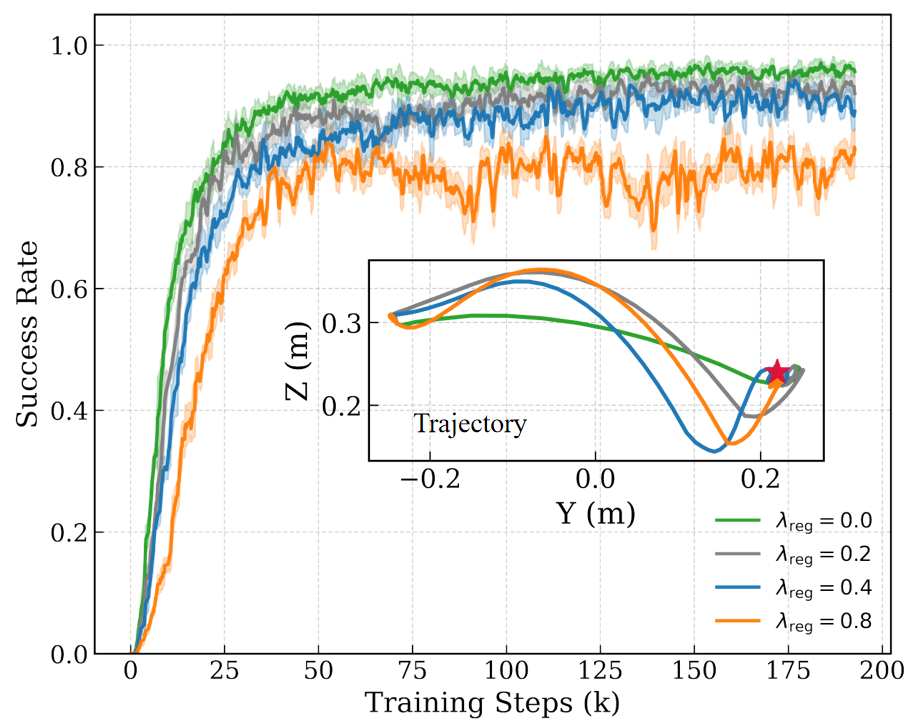}
    \caption{Effect of the regularization weight $\lambda_{\text{reg}}$ on task performance. 
The main plot shows the success rate over training steps, while the inset visualizes representative throwing trajectories under different $\lambda_{\text{reg}}$ values. }
    \label{fig:reg_compare}
\end{figure}

\subsection{Ablation Study}
We ablate the performance of the components introduced in our framework. The success rates during training under different settings are shown in Fig. \ref{fig:ablation}, and the detailed evaluation results are summarized in Table \ref{tab:ablation}. We observe that without human policy regularization, the training achieves a higher success rate upon convergence than all other methods. However, when tested on unseen objects, its success rate drops significantly below that of ours. This indicates that, without incorporating human policy regularization, the policy tends to overfit the training objects and generates unnatural actions, which in turn degrades its generalization capability.
The introduction of hybrid advantage can slightly improve the convergence speed during the early stages of training and enhance policy robustness. Notably, when MARL is not employed (i.e., training directly with PPO), all metrics are the lowest, demonstrating the effectiveness of MARL in learning collaborative tasks. Moreover, compared to single-agent reinforcement learning (SARL), MARL reduces the dimensionality of each agent’s observation space, thereby improving the agent’s robustness.

\subsection{Regularization Coefficient Evaluation}
To evaluate the effect of human policy regularization strength on the agent, we test different regularization weights $\lambda_{\text{reg}}$, with the results shown in Fig. \ref{fig:reg_compare}. Without constraining the thrower with the human policy ($\lambda_{\text{reg}}=0$), the thrower's policy overfits to the training objects, attempting to maximize rewards through unnatural behaviors, and consequently achieves the highest success rate across all $\lambda_{\text{reg}}$ values. As the strength of human policy regularization increases, the policy’s exploration ability gradually decreases, leading to lower training success rates. Therefore, we select $\lambda_{\text{reg}}=0.2$ as the reference value used in the proposed framework. At this value, the policy achieves a balance between constraint and exploration, learning an improved strategy without deviating from the human policy.

\subsection{Baselines Evaluation}

We compare our proposed framework with several works on bimanual dexterous dynamic handover tasks, as summarized in Table~\ref{tab:baseline}. These works cover a wide range of observation modalities, task settings, and algorithmic choices. 
In contrast, our method relies solely on monocular RGB inputs and trains a MAPPO policy to solve abreast catch tasks across multiple object categories. It achieves a success rate of 99\% on training objects and 75\% on unseen objects, outperforming most methods in terms of both accuracy and generalization. These results highlight the effectiveness of our method in learning human-like throwing and catching skills using RGB-based observations, while maintaining robust performance across diverse objects.

\section{CONCLUSION AND FUTURE WORK}

This work introduced DyDexHandover, the first end-to-end RGB-based framework for dual-arm in-air handover. By combining multi-agent reinforcement learning with human-policy regularization, our approach enables robots to perform fluid, human-like throwing and catching without relying on explicit dynamics models or depth sensing. Experiments in Isaac Sim show a 99\% success rate on training objects and 75\% on unseen objects, underscoring the advantages of human-inspired guidance for natural and adaptive dual-arm coordination. These results highlight the promise of leveraging human priors to achieve more robust, generalizable, and human-like collaboration in high-dynamic robotic scenarios.

Considering the dynamic of the environment and the discrepancies between simulation and the real world, these instabilities may pose safety risks during actual deployment. Future work will explore incorporating safety mechanisms into the policy learning process and further enhancing policy generalization to enable transfer to real robots. 

\bibliographystyle{IEEEtran}
\bibliography{IEEEabrv, main}

\begin{thebibliography}{10}
\providecommand{\url}[1]{#1}
\csname url@samestyle\endcsname
\providecommand{\newblock}{\relax}
\providecommand{\bibinfo}[2]{#2}
\providecommand{\BIBentrySTDinterwordspacing}{\spaceskip=0pt\relax}
\providecommand{\BIBentryALTinterwordstretchfactor}{4}
\providecommand{\BIBentryALTinterwordspacing}{\spaceskip=\fontdimen2\font plus
\BIBentryALTinterwordstretchfactor\fontdimen3\font minus \fontdimen4\font\relax}
\providecommand{\BIBforeignlanguage}[2]{{%
\expandafter\ifx\csname l@#1\endcsname\relax
\typeout{** WARNING: IEEEtran.bst: No hyphenation pattern has been}%
\typeout{** loaded for the language `#1'. Using the pattern for}%
\typeout{** the default language instead.}%
\else
\language=\csname l@#1\endcsname
\fi
#2}}
\providecommand{\BIBdecl}{\relax}
\BIBdecl

\bibitem{shao2024bimanual}
Y.~Shao and C.~Xiao, ``Bimanual grasp synthesis for dexterous robot hands,'' \emph{IEEE Robotics and Automation Letters}, 2024.

\bibitem{wang2024learning}
S.~Wang, L.~Sun, M.~Li, P.~Wang, F.~Zha, W.~Guo, and Q.~Li, ``Learning an image-based visual servoing controller for object grasping,'' \emph{International Journal of Humanoid Robotics}, vol.~21, no.~05, p. 2350033, 2024.

\bibitem{zhou2024learning}
B.~Zhou, H.~Yuan, Y.~Fu, and Z.~Lu, ``Learning diverse bimanual dexterous manipulation skills from human demonstrations,'' \emph{arXiv preprint arXiv:2410.02477}, 2024.

\bibitem{huang2023dynamic}
B.~Huang, Y.~Chen, T.~Wang, Y.~Qin, Y.~Yang, N.~Atanasov, and X.~Wang, ``Dynamic handover: Throw and catch with bimanual hands,'' \emph{arXiv preprint arXiv:2309.05655}, 2023.

\bibitem{wang2017generic}
S.~Wang, X.~Wang, B.~Zhan, S.~Wang, and F.~Zha, ``A generic control method of manipulator based on optimization,'' in \emph{2017 2nd International Conference on Advanced Robotics and Mechatronics (ICARM)}.\hskip 1em plus 0.5em minus 0.4em\relax IEEE, 2017, pp. 486--491.

\bibitem{wang2023learning}
S.~Wang, L.~Sun, F.~Zha, W.~Guo, and P.~Wang, ``Learning adaptive reaching and pushing skills using contact information,'' \emph{Frontiers in Neurorobotics}, vol.~17, p. 1271607, 2023.

\bibitem{chen2024object}
Y.~Chen, C.~Wang, Y.~Yang, and C.~K. Liu, ``Object-centric dexterous manipulation from human motion data,'' \emph{arXiv preprint arXiv:2411.04005}, 2024.

\bibitem{wang2022learning}
S.~Wang, W.~Hu, L.~Sun, X.~Wang, and Z.~Li, ``Learning adaptive grasping from human demonstrations,'' \emph{IEEE/ASME Transactions on Mechatronics}, vol.~27, no.~5, pp. 3865--3873, 2022.

\bibitem{zhou2025intelligent}
H.~Zhou and X.~Lin, ``Intelligent redundant manipulation for long-horizon operations with multiple goal-conditioned hierarchical learning,'' \emph{Advanced Robotics}, vol.~39, no.~6, pp. 291--304, 2025.

\bibitem{kober2012learning}
J.~Kober, K.~Muelling, and J.~Peters, ``Learning throwing and catching skills,'' in \emph{2012 IEEE/RSJ International Conference on Intelligent Robots and Systems}.\hskip 1em plus 0.5em minus 0.4em\relax IEEE, 2012, pp. 5167--5168.

\bibitem{chen2022towards}
Y.~Chen, T.~Wu, S.~Wang, X.~Feng, J.~Jiang, Z.~Lu, S.~McAleer, H.~Dong, S.-C. Zhu, and Y.~Yang, ``Towards human-level bimanual dexterous manipulation with reinforcement learning,'' \emph{Advances in Neural Information Processing Systems}, vol.~35, pp. 5150--5163, 2022.

\bibitem{lan2023dexcatch}
F.~Lan, S.~Wang, Y.~Zhang, H.~Xu, O.~Oseni, Z.~Zhang, Y.~Gao, and T.~Zhang, ``Dexcatch: Learning to catch arbitrary objects with dexterous hands,'' \emph{arXiv preprint arXiv:2310.08809}, 2023.

\bibitem{zhan2024safe}
W.~Zhan and P.~Chin, ``Safe multi-agent reinforcement learning for bimanual dexterous manipulation,'' in \emph{2024 IEEE/RSJ International Conference on Intelligent Robots and Systems (IROS)}.\hskip 1em plus 0.5em minus 0.4em\relax IEEE, 2024, pp. 12\,420--12\,427.

\bibitem{yu2022surprising}
C.~Yu, A.~Velu, E.~Vinitsky, J.~Gao, Y.~Wang, A.~Bayen, and Y.~Wu, ``The surprising effectiveness of ppo in cooperative multi-agent games,'' \emph{Advances in neural information processing systems}, vol.~35, pp. 24\,611--24\,624, 2022.

\bibitem{gordon2023somato}
E.~M. Gordon, R.~J. Chauvin, A.~N. Van, A.~Rajesh, A.~Nielsen, D.~J. Newbold, C.~J. Lynch, N.~A. Seider, S.~R. Krimmel, K.~M. Scheidter \emph{et~al.}, ``A somato-cognitive action network alternates with effector regions in motor cortex,'' \emph{Nature}, vol. 617, no. 7960, pp. 351--359, 2023.

\bibitem{fitzpatrick2001primary}
P.~Fitzpatrick, ``The primary motor cortex: upper motor neurons that initiate complex voluntary movements,'' \emph{Neuroscience}, vol.~2, 2001.

\bibitem{ding2020distributed}
G.~Ding, J.~J. Koh, K.~Merckaert, B.~Vanderborght, M.~M. Nicotra, C.~Heckman, A.~Roncone, and L.~Chen, ``Distributed reinforcement learning for cooperative multi-robot object manipulation,'' \emph{arXiv preprint arXiv:2003.09540}, 2020.

\bibitem{liu2021collaborative}
L.~Liu, Q.~Liu, Y.~Song, B.~Pang, X.~Yuan, and Q.~Xu, ``A collaborative control method of dual-arm robots based on deep reinforcement learning,'' \emph{Applied Sciences}, vol.~11, no.~4, p. 1816, 2021.

\bibitem{lin2025sim}
T.~Lin, K.~Sachdev, L.~Fan, J.~Malik, and Y.~Zhu, ``Sim-to-real reinforcement learning for vision-based dexterous manipulation on humanoids,'' \emph{arXiv preprint arXiv:2502.20396}, 2025.

\bibitem{qin2023dexpoint}
Y.~Qin, B.~Huang, Z.-H. Yin, H.~Su, and X.~Wang, ``Dexpoint: Generalizable point cloud reinforcement learning for sim-to-real dexterous manipulation,'' in \emph{Conference on Robot Learning}.\hskip 1em plus 0.5em minus 0.4em\relax PMLR, 2023, pp. 594--605.

\bibitem{wu2023learning}
Y.-H. Wu, J.~Wang, and X.~Wang, ``Learning generalizable dexterous manipulation from human grasp affordance,'' in \emph{Conference on Robot Learning}.\hskip 1em plus 0.5em minus 0.4em\relax PMLR, 2023, pp. 618--629.

\bibitem{van2024geometric}
K.~Van~Wyk, A.~Handa, V.~Makoviychuk, Y.~Guo, A.~Allshire, and N.~D. Ratliff, ``Geometric fabrics: a safe guiding medium for policy learning,'' in \emph{2024 IEEE International Conference on Robotics and Automation (ICRA)}.\hskip 1em plus 0.5em minus 0.4em\relax IEEE, 2024, pp. 6537--6543.

\bibitem{rajeswaran2017learning}
A.~Rajeswaran, V.~Kumar, A.~Gupta, G.~Vezzani, J.~Schulman, E.~Todorov, and S.~Levine, ``Learning complex dexterous manipulation with deep reinforcement learning and demonstrations,'' \emph{arXiv preprint arXiv:1709.10087}, 2017.

\bibitem{wang2024dexcap}
C.~Wang, H.~Shi, W.~Wang, R.~Zhang, L.~Fei-Fei, and C.~K. Liu, ``Dexcap: Scalable and portable mocap data collection system for dexterous manipulation,'' \emph{arXiv preprint arXiv:2403.07788}, 2024.

\bibitem{zhang2025catch}
Y.~Zhang, T.~Liang, Z.~Chen, Y.~Ze, and H.~Xu, ``Catch it! learning to catch in flight with mobile dexterous hands,'' in \emph{2025 IEEE International Conference on Robotics and Automation (ICRA)}.\hskip 1em plus 0.5em minus 0.4em\relax IEEE, 2025, pp. 14\,385--14\,391.

\bibitem{zeng2020tossingbot}
A.~Zeng, S.~Song, J.~Lee, A.~Rodriguez, and T.~Funkhouser, ``Tossingbot: Learning to throw arbitrary objects with residual physics,'' \emph{IEEE Transactions on Robotics}, vol.~36, no.~4, pp. 1307--1319, 2020.

\bibitem{werner2024dynamic}
L.~Werner, F.~Nan, P.~Eyschen, F.~A. Spinelli, H.~Yang, and M.~Hutter, ``Dynamic throwing with robotic material handling machines,'' in \emph{2024 IEEE/RSJ International Conference on Intelligent Robots and Systems (IROS)}.\hskip 1em plus 0.5em minus 0.4em\relax IEEE, 2024, pp. 98--104.

\bibitem{kim2025learning}
T.~Kim, Y.~Yoon, and J.~Kim, ``Learning dexterous bimanual catch skills through adversarial-cooperative heterogeneous-agent reinforcement learning,'' \emph{arXiv preprint arXiv:2502.11437}, 2025.

\bibitem{xiao2021trec}
Y.~Xiao, H.~Yin, S.-H. Wang, and Y.-D. Zhang, ``Trec: Transferred resnet and cbam for detecting brain diseases,'' \emph{Frontiers in Neuroinformatics}, vol.~15, p. 781551, 2021.

\bibitem{zhang2024vn}
H.~Zhang, Y.~Du, S.~Zhao, Y.~Yuan, and Q.~Gao, ``Vn-maddpg: A variable-noise-based multi-agent reinforcement learning algorithm for autonomous vehicles at unsignalized intersections,'' \emph{Electronics}, vol.~13, no.~16, p. 3180, 2024.

\bibitem{mittal2023orbit}
M.~Mittal, C.~Yu, Q.~Yu, J.~Liu, N.~Rudin, D.~Hoeller, J.~L. Yuan, R.~Singh, Y.~Guo, H.~Mazhar \emph{et~al.}, ``Orbit: A unified simulation framework for interactive robot learning environments,'' \emph{IEEE Robotics and Automation Letters}, vol.~8, no.~6, pp. 3740--3747, 2023.

\end{thebibliography}

\end{document}